\documentclass[twocolumn,letterpaper]{IEEEAerospaceCLS}  

\usepackage[]{graphicx}

\usepackage{amsmath,amssymb,amsfonts}
\usepackage{algorithmic}
\usepackage{graphicx}
\usepackage[T1]{fontenc}
\usepackage{hyperref}
\usepackage{textcomp}
\usepackage{subfigure}
\usepackage[ISO]{diffcoeff}
\usepackage{caption}
\usepackage{mathtools}
\usepackage{units}
\usepackage{xcolor}

\newcommand{\ignore}[1]{}  

\begin{document}
\title{Design and Model Predictive Control of a Mars Coaxial Quadrotor}

\author{%
Akash Patel
\and 
Avijit Banerjee
\and 
Bj\"{o}rn Lindqvist 
\and
Christoforos Kanellakis 
\and
George Nikolakopoulos 

\thanks{\footnotesize The authors are with the Robotics and AI Team, Department of Computer, Electrical and Space Engineering, Lule\r{a} University of Technology, Lule\r{a} SE-97187, Sweden}
\thanks{Corresponding Author's email: \texttt{akash.patel@ltu.se} }
}


\maketitle

\thispagestyle{plain}
\pagestyle{plain}

\maketitle

\thispagestyle{plain}
\pagestyle{plain}

\begin{abstract}
Mars has been a prime candidate for planetary exploration of the solar system because of the science discoveries that support chances of future habitation on this planet. The Mars exploration landers and rovers have laid the foundation of our understanding of the planet's atmosphere and terrain. However, the rovers have presented limitations in terms of their pace, traversability, and exploration capabilities from the ground and thus, one of the main field of interest for future robotic mission to Mars is to enhance the autonomy of this exploration vehicles. Martian caves and lava tubes like terrains, which consists of uneven ground, poor visibility and confined space, makes it impossible for wheel based rovers to navigate through these areas. In order to address these limitations and advance the exploration capability in a Martian terrain, this article presents the design and control of a novel coaxial quadrotor Micro Aerial Vehicle (MAV). As it will be presented, the key contributions on the design and control architecture of the proposed Mars coaxial quadrotor, are introducing an alternative and more enhanced, from a control point of view concept, when compared in terms of autonomy to Ingenuity. Based on the presented design, the article will introduce the mathematical modelling and automatic control framework of the vehicle that will consist of a linearised model of a co-axial quadrotor and a corresponding Model Predictive Controller (MPC) for the trajectory tracking. Among the many models, proposed for the aerial flight on Mars, a reliable control architecture lacks in the related state of the art. The MPC based closed loop responses of the proposed MAV will be verified in different conditions during the flight with additional disturbances, induced to replicate a real flight scenario. For the model validation purpose, the Mars coaxial quadrotor is simulated inside a Martian environment with related atmospheric conditions in the Gazebo simulator, which will use the proposed MPC controller for following an a priory defined trajectory. In order to further validate the proposed control architecture and prove the efficacy of the suggested design, the introduced Mars coaxial quadrotor and the MPC scheme will be compared to a PID-type controller, similar to the Ingenuity helicopter's control architecture for the position and the heading.
\end{abstract}

\tableofcontents

\section{Introduction}

The question that has been pursued over many years now is: how did life evolved in our solar system, and other planets from Earth, where else life could have existed in the early days of our solar system formation? These questions have lead us to identifying the habitable zone of our solar system and the planets that fall under this habitable zone. Earth and Mars are prime candidates for satisfying the criteria defined to be in the habitable zone of the solar system. For many years now, we have been sending unmanned robotic probes, orbiters, landers and rovers to the red planet Mars in order to understand its composition, atmosphere, geology etc. There have been over 45 missions that were targeted to Mars by collective effort from different space agencies from around the world. Unmanned robotic missions are gaining more attention nowadays for exploring the red planet, while also surviving the harsh atmosphere environment and radiation conditions. The robotic exploration has paved the way for humanity to research technology levels that supports human settlement on Mars. The rovers Spirit and Opportunity have led a foundation for developing technology to send robust and autonomous systems to conduct experiments on Mars, while orbiters have mapped the surface of the planet with great detail that helps space agencies in selecting the area of interest for future missions. The orbiters like Mars Reconnaissance Orbiter (MRO) and Mars Express have mapped the interesting regions of the planet like Volcanoes, Canyons and polar ice caps. Even after using the best quality Hires Images from MRO, the surface has only been mapped with resolution of about 20 m per pixel. For this purpose, the rovers have been designed and sent to Mars in order to partially fill the data gaps from the orbiters. Furthermore, the rovers have been successful in doing onsite research about mineral composition of rocks and soil, atmospheric studies along with sending beautiful panoramic mosaics of the red planet. The rovers are still limited by their ability to move quickly from one site to another site of interest in order to conduct science experiments.


\subsection{Aerial Vehicle Flight in Mars Atmosphere}

Mars has a very challenging atmosphere in terms of temperature fluctuations, dust storms, surface composition that limits the locations that Mars exploration rovers can reach. The concept of an Aerial Vehicle to fly on Mars is proposed to prove that with considerable optimization in the rotor blade design, enough lift could be generated to fly a lightweight MAV in the thin atmosphere. The concept also focuses on making the flight and operation autonomous in a way such that the vehicle predicts it's next control inputs based on reference trajectory and it's current state. The proposed in this article novel Mars coaxial quadrotor is designed to operate in the conditions where the tip Mach numbers are high and Reynolds number is low. In a generic rotor design, it is important to maintain subsonic speed at the tip of the rotor in order to avoid undesired shock wave formation. Due to the fact that the atmospheric density is very low on Mars, there is a benefit in spinning the rotors at higher RPM and still keeping the tip speeds subsonic. In this approach, the hovering of the vehicle will be controlled in a similar manner to any quadrotor that flies at Earth conditions. The roll, pitch and yaw movement commands will be handled by the proposed MPC controller that will be designed specially to control the co-axial rotors. The proposed size of the rotor blade is 1.12 meters and two rotors spin in opposite directions when mounted in a co-axial manner. This approximation estimates the overall vehicle mass to be around 12 Kilograms, while the detailed designed will be presented in the CAD design section of this article. The larger arms of the Mars coaxial quadrotor are suitable to mount roll out solar arrays on them. These solar arrays can be extended for charging and can be retracted before flying.

\section{State of the Art}

In order to overcome the limitation of the wheeled robots, the idea of aerial exploration has been considered for many years and many methods have been proposed to advance the exploration capability on Mars. In general, the aerial mode of traversing is beneficial in moving from point A to point B with minimal considerations about the terrain and grounded obstacles. Furthermore, aerial exploration also allows for movements at higher speeds from one site to another site of interest. There have been numerous ideas and models that proposed flying or hovering vehicles in other planets. Among different models, studied during the literature study of this article, the Mars helicopter (Ingenuity) and the related Dragonfly Mission to Titan (moon of Saturn) have been a big inspiration for the design and modelling part of this article. The Mars helicopter is a small, lightweight helicopter model that has been designed and tested by JPL at NASA in a simulated environment that represents the atmosphere, density, pressure and gravity in Mars-like conditions. The tech-demo Ingenuity model has also been flown on Mars and has provided promising flight data that validates the idea of autonomous powered flight in Mars atmosphere. The optimized blade span for the mars helicopter is approximately 1.2 meters and the overall vehicle weights about 1.8 Kilograms. 

\begin{figure}[htbp!]
  \centering
    \includegraphics[width=1\linewidth]{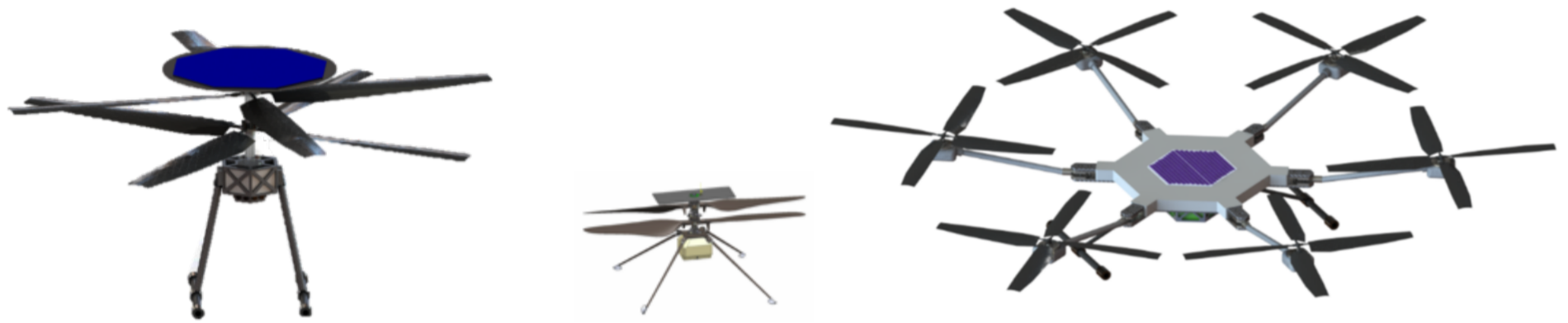}
    \caption{Advanced MSH, Ingenuity and MSH hexacopter}
  \label{fig:MSH}
\end{figure}

The research and development carried out for the Ingenuity helicopter has not only enabled detailed analysis studies in the field of low Reynolds number aerodynamics but also focused on the overall vehicle design. The Ingenuity helicopter is a kind of MAV that works based on the concept of helicopter mechanics. It has two rotors that are mounted co-axially on top of each other. Both rotors spins in opposite directions and this allows the vehicle to balance the torque produced by the rotors. Having co-axial rotors on the Ingenuity helicopter can improve the way a small rotor can produce lift. Because of the co axial system it approximately double the lift produced by the rotor, for the same amount of rotor disk area, and it also helps the turbulent region of the flow to shift downwards. The schematic of the Ingenuity helicopter is shown in \autoref{fig:MSH} (Courtesy of Aerovironment and JPL \cite{pipenberg2019design}). The Ingenuity helicopter is powered by solar arrays that are mounted on the top of the rotor system.

Because of the low overall mass of the system, the required size of solar array in order to power the helicopter is small and therefore the solar arrays are mounted directly on top of the rotor system, while the blades of the Ingenuity helicopter are specifically designed for the thin atmosphere of Mars. In \autoref{fig:MSH} the varying angle of attack for the rotor blades can be seen. The blade profile and its twist distribution is designed in such a way that the free stream velocity increases towards the tip of the rotor blades but the tip mach number still stays subsonic. The flat geometry of the rotor blades can be understood by the role it has to play in order to minimize the free stream velocity and restricting the tip speeds in the subsonic region \cite{pipenberg2019design}.

The control module is responsible for adding feedback control based on the error computed from the reference trajectory and the current state of the vehicle. In this approach, the control inputs are converted into motor speed commands and corresponding actuator commands, which sets the correct pitch angle for the blades~\cite{balaram2018mars}. A detailed control scheme for the Mars helicopter is discussed in~\cite{grip2018guidance}. Because each control loop for the Mars helicopter is designed to work based on the PID controller concept, there are some shortcomings of this control scheme. In terms of stability of the vehicle, the control responses from a PID controller do not yield the best performance, mainly due to the variable gain required to tune PID controller in different operating conditions on Mars. Another limitation of a PID type controller for an autonomous flight on Mars is that there is no prior information about flying conditions available in order to tune the gains for different phase of the flight. Therefore a single set of gains are optimized for the whole control architecture. This lacks in performance when flying in disturbed airflow on Mars. Other than the Mars helicopter, there have been a few more rotor craft based vehicle designs proposed for Mars exploration as discussed in~\cite{radotich2021study}. Among the designs discussed in~\cite{radotich2021study}, the Advanced Mars Science Helicopter and hexacopter variant of Mars Science Helicopter (MSH) also presents a similar design considerations and the approach is adapted for our Mars coaxial quadrotor. The designs of the Advanced MSH, Ingenuity and MSH-hexacopter respectively are shown in \autoref{fig:MSH} \cite{johnson2020mars} for a general size comparison.

The coaxial rotor system of the advanced MSH is derived from the Ingenuity's design. However, further modifications are proposed in~\cite{withrow2020advanced}. The aim of the MSH-hexacopter proposed in~\cite{johnson2020mars} is to have multi rotor system that is able to carry more science payload, as well as an enhanced time of flight for the vehicle. It should be also noted that having redundancy in a multi rotor (MSH-hexacopter) system is a great advantage, while aiming at an autonomous flight on another planet. As discussed previously, the design aspect of rotor crafts for mars are comprehensively mentioned in different literature but an advanced control architecture is lacking in many of them. The greater goal and the major contribution of this article is to address the control aspect of the proposed rotor craft and validate it on a Mars like simulation environment.

\section{Contributions}
Based on the above mentioned state-of-the-art, the main contributions of this article are listed in the following manner.

The first contribution to this work is to design a multi rotor MAV that can accommodate a large coaxial rotor system. The design of the coaxial rotor system is derived from the state of the art. The proposed design allows to use the concept of roll out solar arrays for powering the vehicle. Even though the rotor span is considerably high, the large arms of the coaxial quadrotor benefit in accommodating the subsystem that rolls out solar arrays for charging when the vehicle is stationary. This allows for a large area of solar arrays, which directly result into more power produced for charging a bigger battery faster, while the overall aim was to increase the flight duration by optimally designing the structure of the vehicle.

In addition to the contribution in the design of the vehicle, 3D flow simulations are carried out in the Fluent module of the Ansys software. In order to compute the thrust, produced by the system of co axial rotors, the pressure on top and bottom rotor is calculated for certain RPM at the Mars atmosphere reference conditions, as mentioned in \autoref{table:1}. The thrust curve for one set of the coaxial rotor system shows that the thrust value saturates after certain time into the simulation and the saturation value is in acceptable range for the respective size and rotation speed of the rotor system. The 3D flow simulations also validates that the coaxial system prevents the shock wave formation at the tip of the rotors when operating at certain RPM. 

The main contribution in the modelling and control part of this article is to design a non linear mathematical model of a coaxial quadrotor and developing a PID, as well as a MPC controller for simulating the control of the vehicle. The design of the motor mixing is different from the conventional quadrotor because of the eight rotors mounted co-axially in the proposed coaxial quadrotor configuration. Open loop simulations are carried out before designing the controller in order to validate the mathematical model of the coaxial quadrotor. The loop is closed with the help of the Model Predictive Control, while in the proposed mathematical model, because of the highly non linear equations, the model needed to be linearised around a reference trajectory in order to control the vehicle using the linear MPC framework. The control responses from the MPC are compared against a simple PID controller for the same model and in the results as it will be presented it is evident that the MPC based model outperforms the state of the art PID control based model.

\section{Design and Modelling of Mars coaxial quadrotor}

\begin{figure}[h!]
    \centering
    \subfigure[Mars coaxial quadrotor in flight mode]
    {
        \includegraphics[width=1\linewidth]{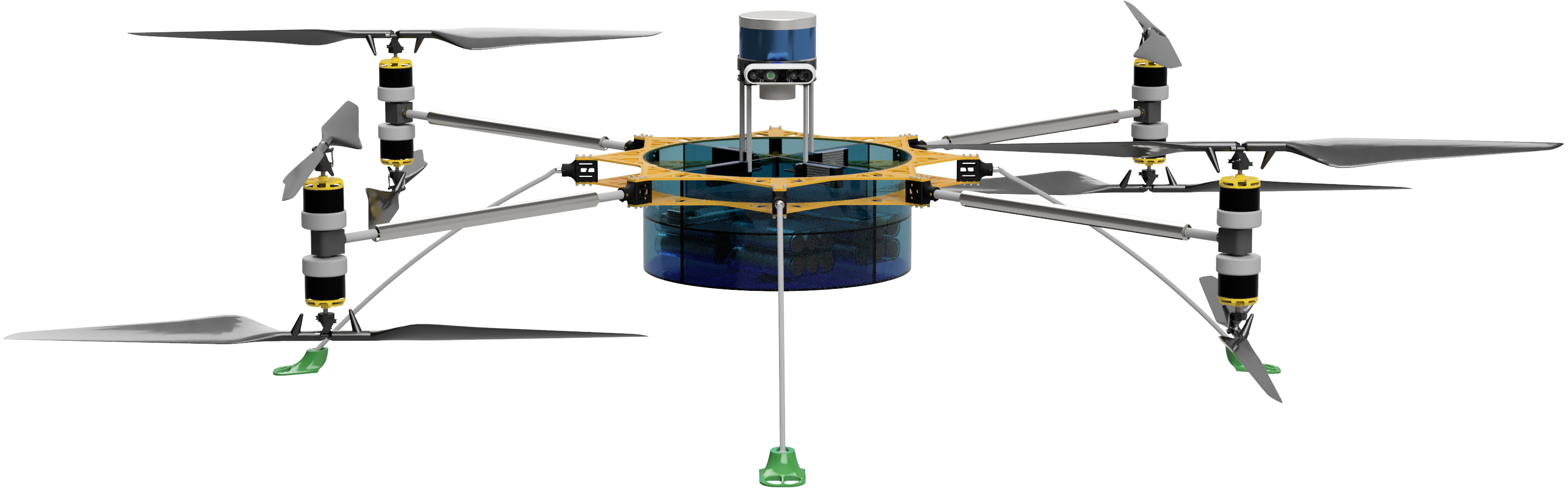}
        \label{1}
    }
    \subfigure[Mars coaxial quadrotor in idle mode]
    {
        \includegraphics[width=1\linewidth]{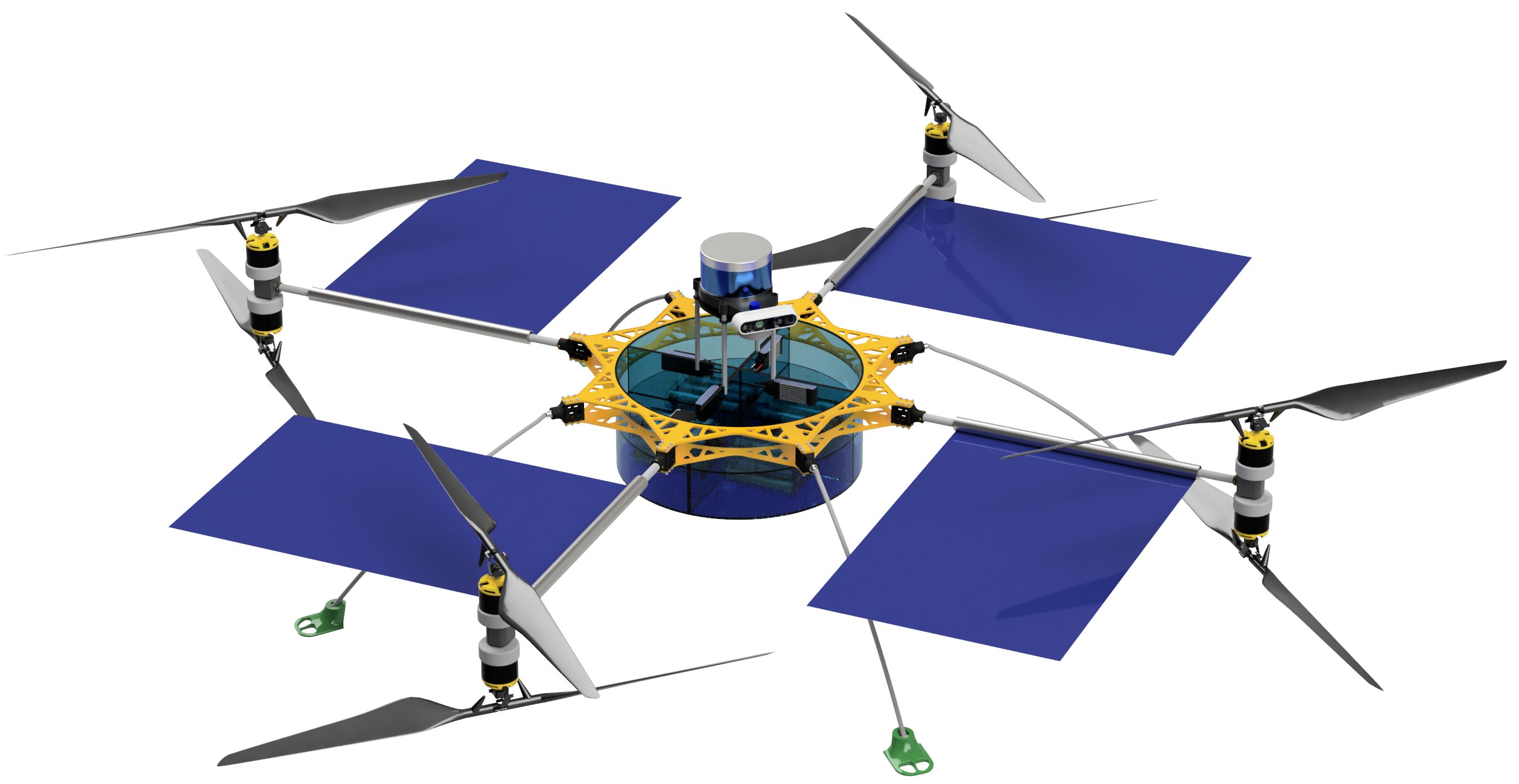}
        \label{2}
    }
    \caption{Mars coaxial quadrotor CAD model}
    \label{cad}
\end{figure}

The proposed Mars coaxial quadrotor CAD model is designed with a reference to the diameter of the rotor blade, optimized fora  sufficient thrust in Mars atmosphere conditions. The CAD model is presented in Figure \autoref{cad}. The proposed design allows for a novel, in terms of accommodating, bigger roll out solar arrays (ROSA)~\cite{7500723} and increased payload capacity. The roll out solar arrays are mounted on the arms of the vehicle and can be extended for charging when the vehicle is in idle mode on ground. Because the roll out solar arrays can be retracted back in a spiral shape that wounds around the arm of the vehicle, it is possible to accommodate a larger area of the solar arrays. This allows for a larger battery capacity, hence larger payload space, because the vehicle can produce significantly high thrust compared to the state of the art. 

The proposed model can be either in idle mode or in flight mode. In idle mode, the vehicle will extend its solar arrays for charging the batteries as shown in~\autoref{2} and when the batteries are charged, the solar arrays can be retracted back for preparation of next flight as shown in~\autoref{1}. Furthermore, the vehicle is equipped with a set of sensors (3D liadr and stereo camera), which help in navigation and obstacle avoidance. 

\subsection{Design Optimizations for mars conditions}

There are two goals that needs to be addressed through the fluid simulations. The first goal is to extract the thrust for the system of the coaxial rotors. The second aim is to observe the flow behaviour around the solid body of the rotor blades. The interaction of flow lines with the solid body gives insights into the transition phase of the flow at different CFD stations (CFD stations are considered as locations from the hub of the rotor, out in the radial direction towards the tip of the rotor). The analysis is also done at a number of different rotation speeds to observe the turbulence nature of the flow when the revolutions are increased. The thrust is calculated as the pressure force, applied by the fluid on to the rotor solid body surfaces. The total thrust that will be displayed in the plots is the sum of the thrust force on upper rotor and the lower rotor. The conditions used for the simulation in a Mars environment are as mentioned in Table \ref{table:1}.

\begin{table}[h!]
\centering
\begin{tabular}{||c c c||} 
 \hline
 Parameter & MARS & EARTH \\ [0.5ex] 
 \hline\hline
 Density, $\rho$ (Kg/$m^3$) & 0.017 & 1.225 \\ 
 Static Pressure \textit{p} (Pa) & 720 & 101325 \\
 Temperature \textit{T} (K) & 223 & 288.20 \\
 Gas constant \textit{R} ($m^{2}/s^{2}$/K) & 188.90 & 287.10 \\
 Dynamic viscosity \textit{$\mu$} (Ns/$m^2$) & 1.130$\cdot$$10^{-5}$ & 1.175$\cdot$$10^{-5}$\\
 Gamma \textit{$\gamma$} & 1.289 & 1.4 \\ [1ex] 
 \hline
\end{tabular}
\caption{Mars and Earth atmospheric conditions}
\label{table:1}
\end{table}

The proposed design of Mars coaxial quadrotor uses four pairs of rotors. The idea of proposing a design with coaxial rotors is validated through CFD simulations in this Section. For this purpose, the first set of simulation is carried out with a single rotor, spinning at 3200 rpm. The results clearly show the effect of the high rpm considered for the desired thrust from a single rotor.

\begin{figure}[h!]
    \centering
    \subfigure[Thrust plot for a single rotor at 3200 rpm]
    {
        \includegraphics[width=1\linewidth]{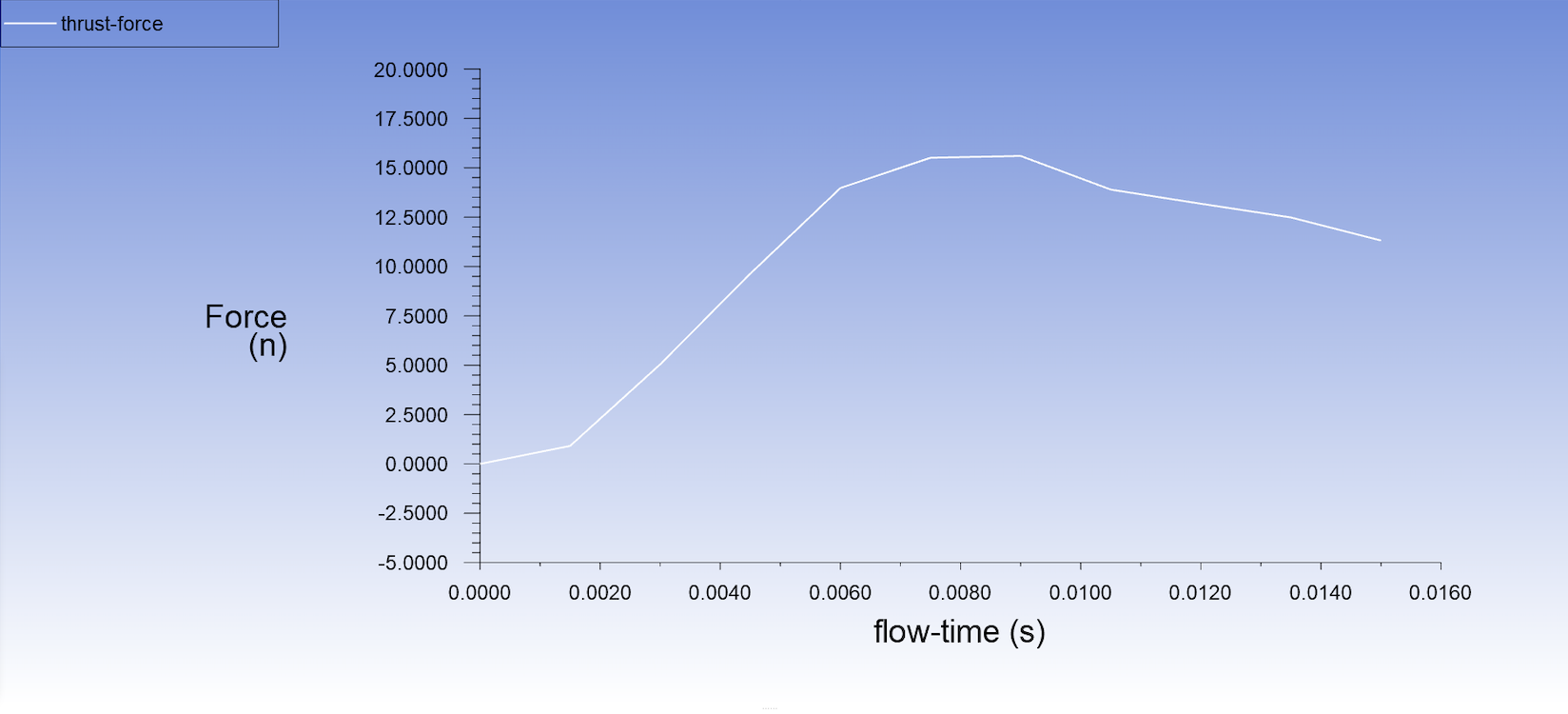}
        \label{thrustplot}
    }
    \subfigure[Velocity contour for a single rotor]
    {
        \includegraphics[width=1\linewidth]{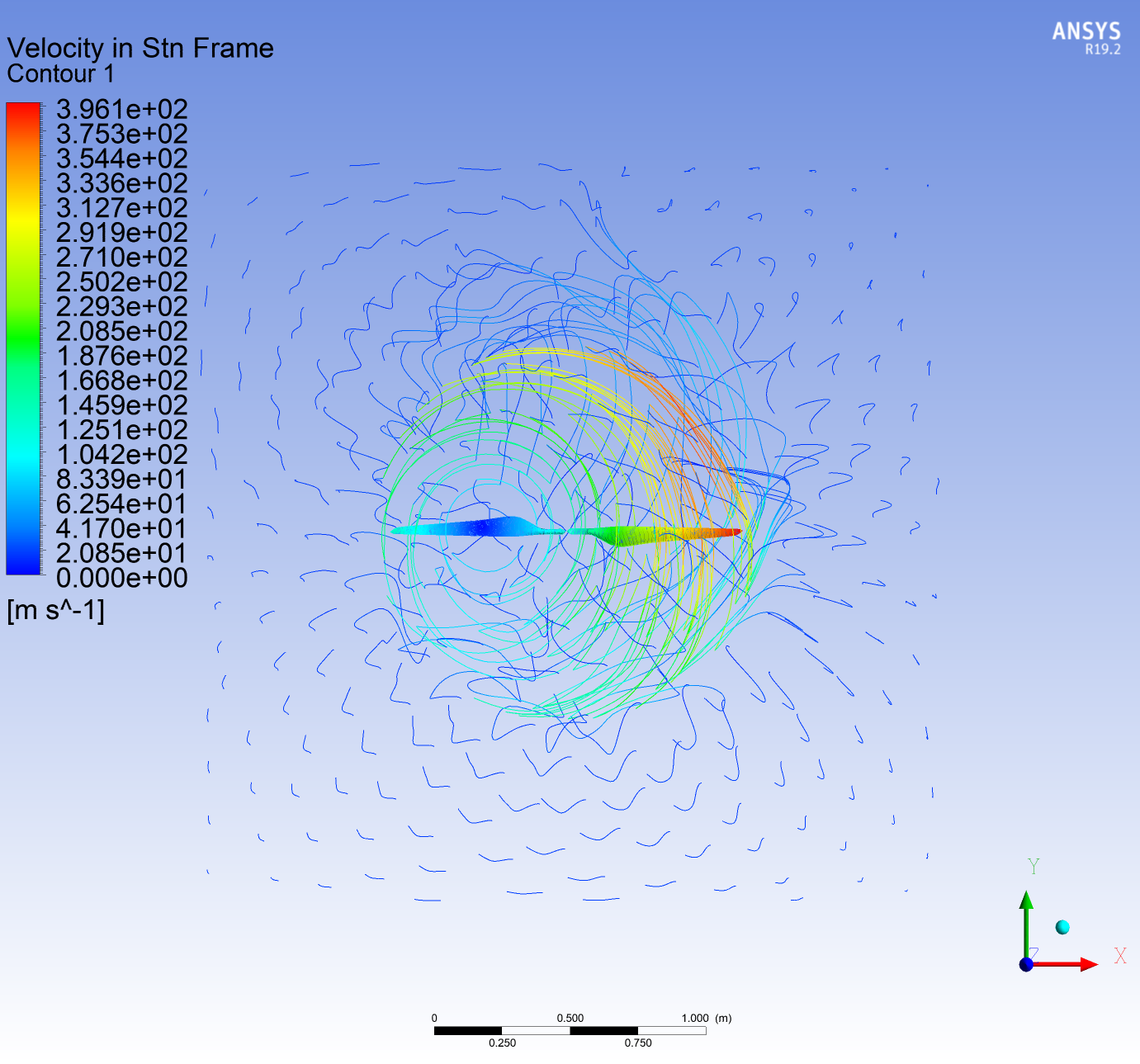}
        \label{velcont}
    }
    \caption{Single rotor flow behaviour}
\end{figure}

In~\autoref{thrustplot} the thrust force is plotted against the flow run time. After the flow convergence, the thrust value at 3200 rpm is observed to be around 11.5 Newtons. The desired thrust value is high when considering the low atmospheric density and the surface pressure used for the simulation (Mars conditions). On the other hand, in order to reach this thrust value, we have provided a high rpm value to the rotor. The effect of the high rpm for a rotor of this size is observed in~\autoref{velcont}. The turbulent nature of the flow and the related disturbances in the stream lines is easily seen towards the tip of the rotor. The speed of sound is given by $a = \sqrt{\gamma R T}$.

For the average surface temperature on Mars, the speed of sound is computed to be around 244 m/s. In the~\autoref{velcont} the velocity in a stationary frame is observed to be around 396 m/s at the tip of the rotor, which results in a mach number around 1.62 and this is clearly not subsonic flow. Thus, the turbulent behaviour of the flow around the tip is the influence from the shock wave that is generated at the tip of the rotor blade. Thus, it can be concluded that the high value of thrust can be achieved by increasing the rpm but at the cost of having a supersonic flow. Also, once the flow becomes supersonic, the thrust decreases exponentially until the RPM is not lowered. This effect would limit the operation of a MAV on Mars. For the co-axial rotor simulation, both rotors are at zero angle of rotation at the beginning of the simulation. The presented set of simulations were carried out at 2800 rpm. The CFD data, extracted from the solver setup, was post processed in the CFD post processor of Fluent and the different contours of velocity, pressure and turbulence kinetic energy was plotted in graphically readable manner.

\begin{figure}[htbp!]
  \centering
  \captionsetup{justification=centering}
    \includegraphics[width=1\linewidth]{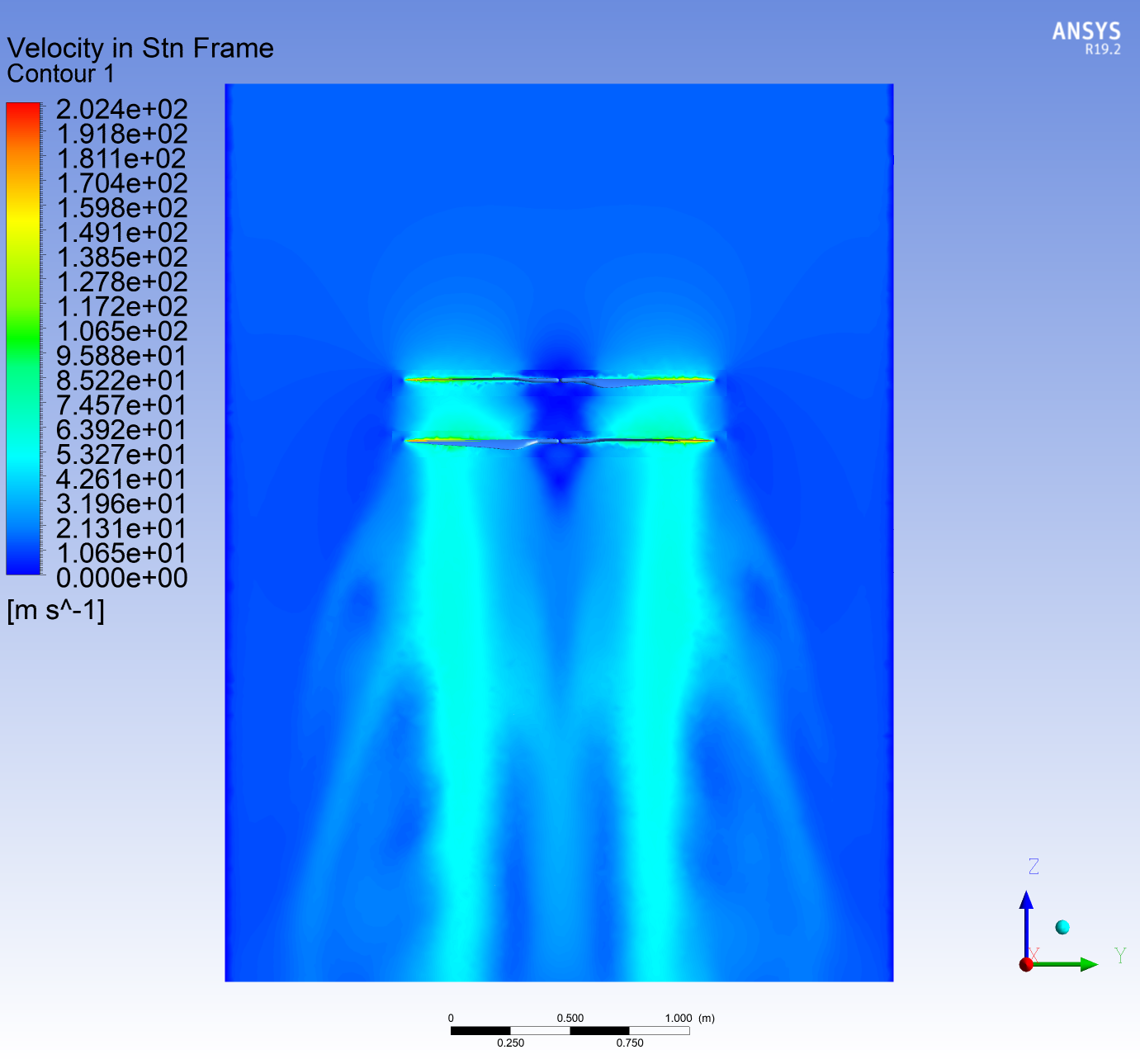}
    \caption{Velocity contour for co axial rotors}
  \label{fig:velinstatn}
\end{figure}

The co axial configuration allows to decrease the revolutions per minute considered for the rotors, while at the same time increasing the overall thrust of the system. In~\autoref{fig:velinstatn} it can be observed that the maximum velocity value at any point inside the considered bounding box is not going beyond 202.4 m/s. This means that even at the tips of the rotor blade, the velocity is limited below the 202.4 m/s value. This implies that the mach number in this case is less than 0.82. Therefore, the flow is always in the subsonic region and the shock waves do not form in this case. The wake is generated in this case because of the influence of the upper rotor on the flow around the lower rotor but the effect of wake is not prominent in this case because due to the optimized shape of the rotor.

\begin{figure}[htbp!]
  \centering
  \captionsetup{justification=centering}
    \includegraphics[width=1\linewidth]{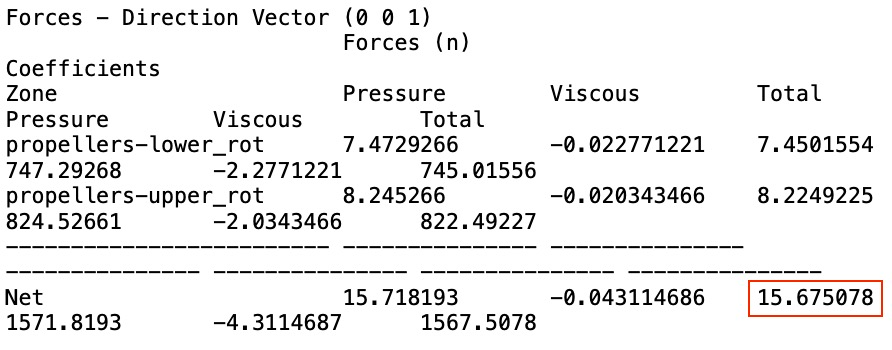}\\
    \caption{Force report from Fluent}
  \label{fig:forcereport}
\end{figure}

The \autoref{fig:forcereport} shows the force report from Fluent including the pressure and viscous forces. The dominant force is the pressure force, which is responsible for the thrust. The force is calculated on the upper and lower rotor separately based on the Mars conditions and rotor's geometry. The total force is around 15.67 Newtons and that is the thrust force for single pair of rotors in the coaxial configuration at 2800 rpm.

\begin{figure}[htbp!]
  \centering
  \captionsetup{justification=centering}
    \includegraphics[width=1\linewidth]{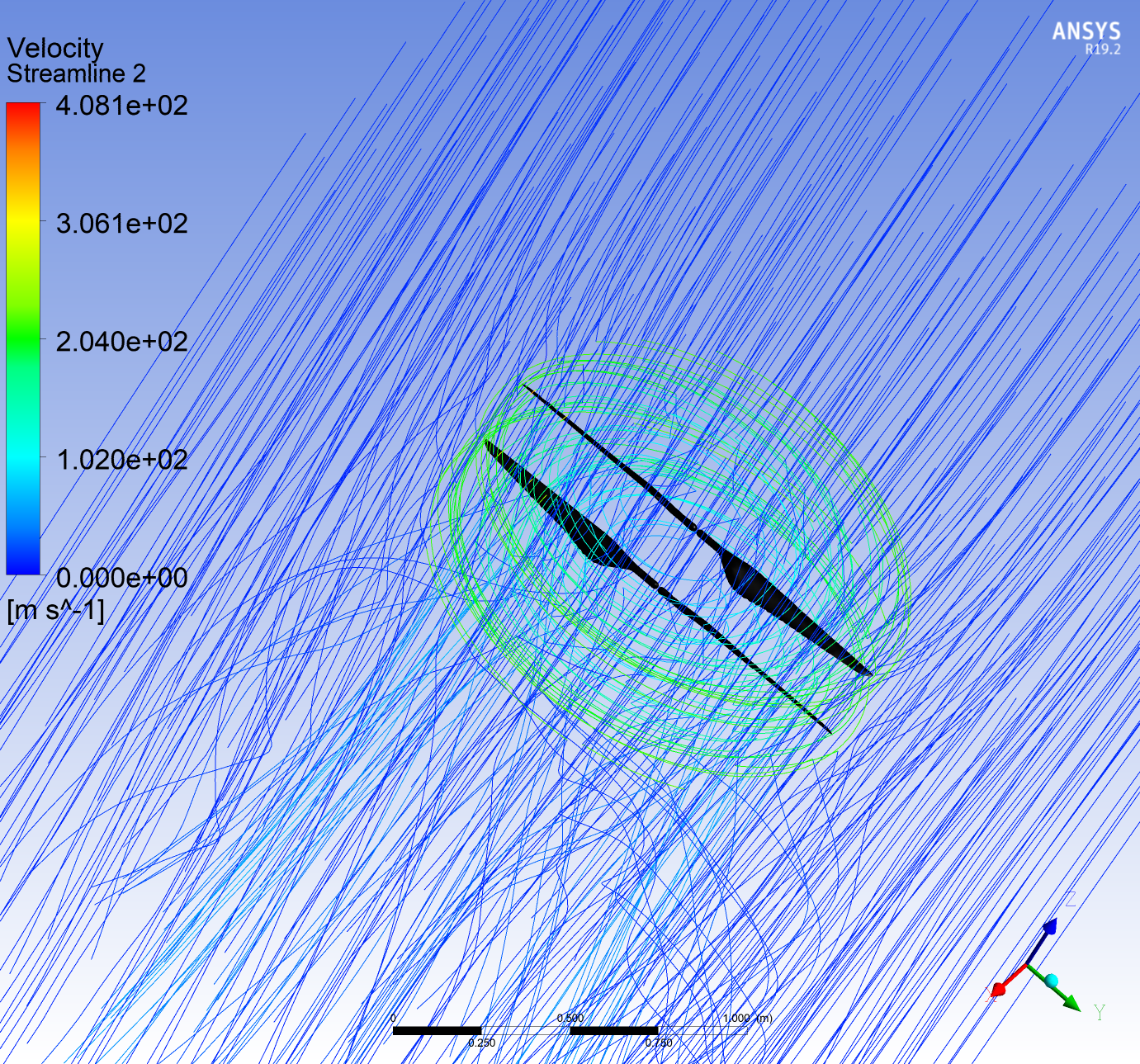}
    \caption{Global velocity contour with streamlines}
  \label{fig:streamline}
\end{figure}

In order to visualize the effect of the turbulence inside the control volume, the flow is also also plotted in terms of streamline and their velocity. The work in~\autoref{fig:streamline} shows the streamlines flowing in from the positive $Z$ direction. as compared to the results shown in~\autoref{velcont}, the streamline behaviour supports the absence of a shock at the tip of the rotor. The rotors are displayed in black to enhance the visualization of flow behaviour around the rotor and inside the rotating module of control volume.

\section{Control Oriented Modelling}

\begin{figure}[h!]
  \centering
  \captionsetup{justification=centering}
    \includegraphics[width=\linewidth]{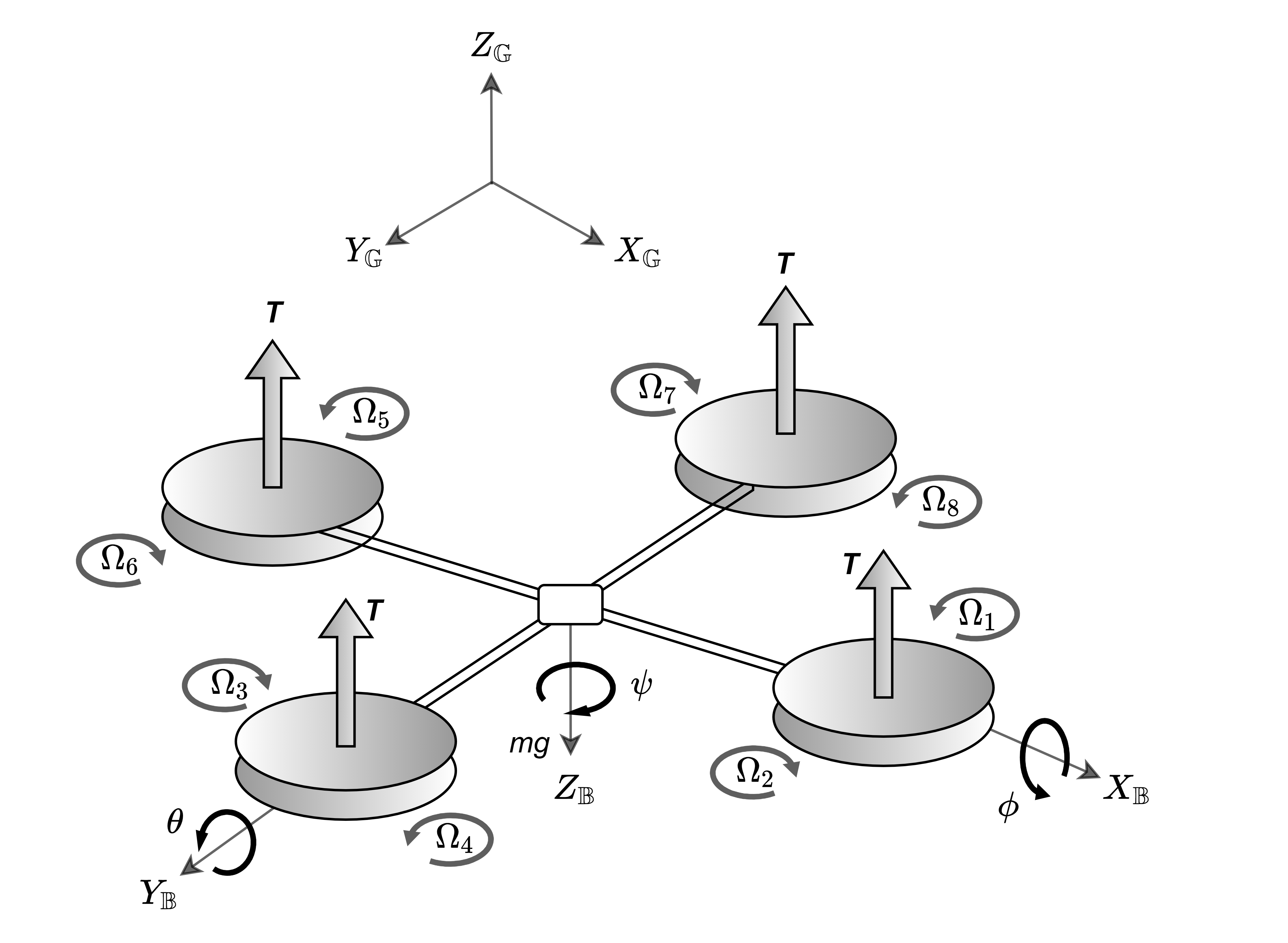}
    \caption{Mars coaxial rotors representation}
  \label{fig:coaxial}
\end{figure}

In order to define the equations of motion of a quadrotor, it is important to define the notations and frame of reference first. As shown in~\autoref{fig:coaxial}, two frames of reference are considered. A ground frame of reference, which is always stationary and the other is the quadrotor body frame of reference. The structure and motion of the vehicle are described in the body frame of reference, where the position commands are given in the ground frame of reference. In the mathematical model, the position values will be rotated by the heading angle $\psi$ in order to utilize them for roll/pitch calculation in the body frame.

The proposed Mars coaxial quadrotor is a coaxial MAV that is inspired from the conventional quadrotor design. The goal is to produce more thrust by using a coaxial system of rotors. The quadrotor's overall control is achieved by providing input to the four motor speeds, whereas in the Mars coaxial quadrotor system, the control will be achieved by providing input to the eight motor speeds. The direction of rotation for each rotor for the Mars coaxial quadrotor is as depicted in~\autoref{fig:coaxial}.

In order to balance the total torque of the vehicle, each pair of coaxial rotors spin in opposite direction with respect to the adjacent pair of rotors. Since the rotor is in the $+$ configuration, the forward direction is the $X$ axis on the body frame. The rotor speed is denoted as $\Omega$ and the total thrust in the upward direction is denoted as $T$. Therefore,

\begin{equation*}
    T_{i} = K_{T}\Omega_{i}^{2},\  i = 1,2,..7,8
\end{equation*}

$g = 3.711 m/s^2$ is the gravitational acceleration on Mars, $B$ is the aerodynamic friction coefficient and $T\ =\ \sum T_{i}$ is the total thrust in the body frame. Considering the Newton's second law of motion in the inertial coordinate frame:

\begin{equation}
    m\dot{V} = 
    \begin{pmatrix}
    0 \\ 0 \\ mg
    \end{pmatrix} -R
    \begin{pmatrix}
    0 \\ 0 \\ T
    \end{pmatrix} - BV
\end{equation}

The distance from the rotor hub to the center of gravity of the vehicle is $d$, therefore the roll moment of the vehicle about the $X$ axis is given as: 

\begin{equation}
    U_2 = d K_{T}(\Omega_{7}^{2} + \Omega_{8}^{2} - \Omega_{3}^{2} - \Omega_{4}^{2})
\end{equation}

The pitch moment of the vehicle about the $Y$ axis is given as:

\begin{equation}
    U_3 = d K_{T}(\Omega_{5}^{2} + \Omega_{6}^{2} - \Omega_{1}^{2} - \Omega_{2}^{2})
\end{equation}

If the coefficient of drag is $K_{D}$ then, the yaw moment of the vehicle about $Z$ axis is given as:

\begin{equation}
    U_4 = K_{D}(\Omega_{2}^{2} + \Omega_{4}^{2} + \Omega_{6}^{2} + \Omega_{8}^{2} - \Omega_{1}^{2} - \Omega_{3}^{2} - \Omega_{5}^{2} - \Omega_{7}^{2})
\end{equation}

The total thrust $T$ is represented as $U1$ for the throttle control in the Matlab model. The overall rotor speed is represented as:

\begin{equation}
    \Omega_{r} = - \Omega_{1} + \Omega_{2} - \Omega_{3} + \Omega_{4} - \Omega_{5} + \Omega_{6} - \Omega_{7} + \Omega_{8}
\end{equation}

The total rotational accleration of the vehicle is described as:

\begin{equation}
    I\dot{\omega} = -\omega \times I\omega + \tau
\end{equation}

Here $\omega$ is the angular velocity vector in inertial frame of reference. If only the thrust force is considered in the inertial frame of reference then the total forces and the torque matrix is given as:
\small
\begin{equation*}
    \begin{psmallmatrix}
    T \\ \tau
    \end{psmallmatrix} =
    \begin{psmallmatrix}
    -K_{T}& -K_{T}& -K_{T}& -K_{T}& -K_{T}& -K_{T}& -K_{T}& -K_{T} \\ 0& 0& -dK_{T}& -dK_{T}& 0& 0& dK_{T}& dK_{T} \\ -dK_{T}& -dK_{T}& 0& 0& dK_{T}& dK_{T}& 0& 0 \\ -K_{D}& K_{D}& -K_{D}& K_{D}& -K_{D}& K_{D}& -K_{D}& K_{D}
    \end{psmallmatrix}
    \begin{psmallmatrix}
    \Omega_{1}^{2} \\ \Omega_{2}^{2} \\ \Omega_{3}^{2} \\ \Omega_{4}^{2} \\ \Omega_{5}^{2} \\ \Omega_{6}^{2} \\ \Omega_{7}^{2} \\ \Omega_{8}^{2}
    \end{psmallmatrix} 
 \end{equation*}
\begin{equation*}
    = A 
    [\Omega_{1}^{2}, \Omega_{2}^{2},..,\Omega_{8}^{2}]^T
\end{equation*}

\normalsize
Because the thrust coefficient $K_{T}$, the drag coefficient $K_{D}$ and the distance from center of gravity $d$ are constant and positive values, the rotor speeds for the input can be computed from the thrust and torque required values. Therefore, the rotor speeds are formulated as shown in equation (\ref{speeds}):

\begin{equation}
    [\Omega_{1}^{2}, \Omega_{2}^{2},..,\Omega_{8}^{2}]^T = A^{-1}
    \begin{psmallmatrix}
    T \\ \tau_{x} \\ \tau_{y} \\ \tau_{z}
    \end{psmallmatrix}
    \label{speeds}
\end{equation}

\subsection{Nonlinear Model}

Considering that the assumptions made for the quadrotor also hold for the Mars coaxial quadrotor, the equations of motion can be formulated using the Newton-Euler concept. The origin of the body frame of reference coincides with the center of gravity of the vehicle, while the linear acceleration in the $x$ direction is given as:

\begin{equation}
    \Ddot{x} = [\sin{\psi}\sin{\phi} + \cos{\psi}\sin{\theta}\cos{\phi}]\frac{U_{1}}{m}
    \label{xdotdot}
\end{equation}

The linear acceleration in the $y$ direction is given as:

\begin{equation}
    \Ddot{y} = [-\cos{\psi}\sin{\phi} + \sin{\psi}\sin{\theta}\cos{\phi}]\frac{U_{1}}{m}
\end{equation}

and the linear acceleration in the $z$ direction is given as:

\begin{equation}
    \Ddot{z} = [\cos{\theta}\cos{\phi}]\frac{U_{1}}{m} - g
\end{equation}

Similarly, the rotational dynamics can be described by the angular acceleration in the roll, pitch and yaw directions. The angular accelerations are given as:

\begin{equation}
    \Ddot{\phi} = [ \dot{\theta} \dot{\psi} (I_{yy} - I_{zz}) - J_{r} \dot{\theta} \Omega_{r} + U_{2}]\frac{1}{I_{xx}}
    \label{phidotdot}
\end{equation}

\begin{equation}
    \Ddot{\theta} = [ \dot{\phi} \dot{\psi} (I_{zz} - I_{xx}) - J_{r} \dot{\phi} \Omega_{r} + U_{3}]\frac{1}{I_{yy}}
    \label{thetadotdot}
\end{equation}

\begin{equation}
    \Ddot{\psi} = [ \dot{\phi} \dot{\theta} (I_{xx} - I_{yy}) + U_{4}]\frac{1}{I_{zz}}
    \label{psidotdot}
\end{equation}

In equations (\ref{phidotdot}) and (\ref{thetadotdot}), $J_{r}$ is the total rotational inertia of the vehicle.

In order to find the state space model, the system of nonlinear equations can be written in the form of, \\ $ X = [x,\ y,\ z,\ v_{x},\ v_{y},\ v_{z},\ \phi,\  \theta,\  \psi,\ \dot{\phi},\ \dot{\theta},\ \dot{\psi}]^{T}$ 

\noindent The equation (\ref{statespace}) represents the overall state space representation of the Mars coaxial quadrotor mathematical model. 

\begin{equation}
    \dot{X} = 
    \begin{bmatrix}
    v_{x} \\ [\sin{\psi}\sin{\phi} + \cos{\psi}\sin{\theta}\cos{\phi}]\frac{U_{1}}{m} \\ v_{y} \\ [-\cos{\psi}\sin{\phi} + \sin{\psi}\sin{\theta}\cos{\phi}]\frac{U_{1}}{m} \\ v_{z}  \\  [\cos{\theta}\cos{\phi}]\frac{U_{1}}{m} - g \\ \dot{\phi} \\ [ \dot{\theta} \dot{\psi} (I_{yy} - I_{zz}) - J_{r} \dot{\theta} \Omega_{r} + U_{2}]\frac{1}{I_{xx}} \\ \dot{\theta} \\ [ \dot{\phi} \dot{\psi} (I_{zz} - I_{xx}) - J_{r} \dot{\phi} \Omega_{r} + U_{3}]\frac{1}{I_{yy}} \\ \dot{\psi} \\ [ \dot{\phi} \dot{\theta} (I_{xx} - I_{yy}) + U_{4}]\frac{1}{I_{zz}}
    \end{bmatrix}
    \label{statespace}
\end{equation}

\subsubsection{Linearised Model}

The previously defined model of the Mars coaxial quadrotor in equations (\ref{xdotdot}) to (\ref{psidotdot}) is highly nonlinear. In order to design a linear Model Predictive Controller for the system, the model needs to be linearised around a reference point. In this case, the reference point takes as near at the hover condition in equilibrium. Therefore, the following small angle approximations are made for near hover condition. 

\begin{equation*}
    \dot{\phi} \cong \dot{\theta} \cong \dot{\psi} \cong 0
\end{equation*}
\begin{equation*}
    \sin{\phi} \cong \phi\ ;\ \sin{\theta} \cong \theta\ ;\ \sin{\psi} \cong 0
\end{equation*}

In the near hover condition, the Yaw angle $\psi$ and the Yaw rate $\dot{\psi}$ are considered to be zero. Therefore, the position in $x$ and $y$ are solely dependent on $\theta$ and $\phi$ respectively. Considering the near hover condition, the nonlinear model of the Mars coaxial quadrotor can be simplified as:

\begin{equation*}
    \Ddot{x} = g\theta ; \ \Ddot{y} = -\ g\phi ; \ \Ddot{z} = \frac{U_{1}}{m}
\end{equation*}
\begin{equation*}
    \Ddot{\phi} = \frac{U_{2}}{I_{xx}} ; \ \Ddot{\theta} = \frac{U_{3}}{I_{yy}} ; \ \Ddot{\psi} = \frac{U_{4}}{I_{zz}}
\end{equation*}

The State Space representation is a mathematical model of the Mars coaxial quadrotor system as a set of input, output and state variables that are related by first order differential equations. In the state space representation, the state space is the space that consists of all the state variables as its axis. The generic state space representation of the linear system is given as:
\begin{align} \label{ss1}
    \dot{X}(t) = Ax(t) + Bu(t)\\\label{ss2}
     Y(t) = Cx(t) + Du(t)
    \end{align}
In equations (\ref{ss1}) and (\ref{ss2}), 
\noindent $x(t)$ is the 'State Vector',
$y(t)$ is the 'Output Vector', 
$u(t)$ is the 'Input Vector', 
$A$ is the 'System Matrix', 
$B$ is the 'Input Matrix', 
$C$ is the 'Output Matrix' and  
$D$ is the 'Feed forward Matrix'.  
The vector $[x\ y\ z\ \phi\ \theta\ \psi]$ contains the linear and angular positions of the Mars coaxial quadrotor in the round frame of reference. Whereas, the vector $[\dot{x}\ \dot{y}\ \dot{z}\ \dot{\phi}\ \dot{\theta}\ \dot{\psi}]$ contains the linear and angular velocities in the body fixed frame of reference. The linearised state space model can be derived at the equilibrium condition, where $x = x_{0}$ and $u = u_{0}$. $x_{0}$ and $u_{0}$ represent the reference values at the near hover condition. The mathematical model of the Mars coaxial quadrotor used for this article is an actuator based model. In order to make the system more realistic, the control inputs for the proposed mathematical model are the rotor speeds, instead of the corresponding force or torque. In order to linearize the proposed model around the near hover point, the $A$ and $B$ matrices for the state space representation are derived below.

\begin{equation*}
    A = \diffp {F(X,U)}x[x_{0},u_{0}]\ ;\  B = \diffp {F(X,U)}u[x_{0},u_{0}]
\end{equation*}

Thus, the state space equations for the Mars coaxial quadrotor system are mentioned below. 
\small
\begin{eqnarray*}
    \overbrace{
    \begin{bsmallmatrix}
    \dot{x} \\ \Ddot{x} \\ \dot{y} \\ \Ddot{y} \\ \dot{z} \\ \Ddot{z} \\ \dot{\phi} \\ \Ddot{\phi} \\ \dot{\theta} \\ \Ddot{\theta} \\ \dot{\psi} \\ \Ddot{\psi}
    \end{bsmallmatrix}}^\text{$\dot{X}(t)$}
    = 
    \overbrace{
    \begin{bsmallmatrix}
    0 & 1 & 0 & 0 & 0 & 0 & 0 & 0 & 0 & 0 & 0 & 0 \\ 0 & 0 & 0 & 0 & 0 & 0 & 0 & 0 & g & 0 & 0 & 0 \\ 0 & 0 & 0 & 1 & 0 & 0 & 0 & 0 & 0 & 0 & 0 & 0 \\ 0 & 0 & 0 & 0 & 0 & 0 & -g & 0 & 0 & 0 & 0 & 0 \\ 0 & 0 & 0 & 0 & 0 & 1 & 0 & 0 & 0 & 0 & 0 & 0 \\ 0 & 0 & 0 & 0 & 0 & 0 & 0 & 0 & 0 & 0 & 0 & 0 \\ 0 & 0 & 0 & 0 & 0 & 0 & 0 & 1 & 0 & 0 & 0 & 0 \\ 0 & 0 & 0 & 0 & 0 & 0 & 0 & 0 & 0 & 0 & 0 & 0 \\ 0 & 0 & 0 & 0 & 0 & 0 & 0 & 0 & 0 & 1 & 0 & 0 \\ 0 & 0 & 0 & 0 & 0 & 0 & 0 & 0 & 0 & 0 & 0 & 0 \\ 0 & 0 & 0 & 0 & 0 & 0 & 0 & 0 & 0 & 0 & 0 & 1 \\ 0 & 0 & 0 & 0 & 0 & 0 & 0 & 0 & 0 & 0 & 0 & 0 
    \end{bsmallmatrix}}^\text{A}
    \overbrace{
    \begin{bsmallmatrix}
    x \\ \dot{x} \\ y \\ \dot{y} \\ z \\ \dot{z} \\ \phi \\ \dot{\phi} \\ \theta \\ \dot{\theta} \\ \psi \\ \dot{\psi}
    \end{bsmallmatrix}}^\text{$x(t)$}
    + \\
    \overbrace{
    \begin{bsmallmatrix}
    0 & 0 & 0 & 0 & 0 & 0 & 0 & 0 \\ 0 & 0 & 0 & 0 & 0 & 0 & 0 & 0 \\ 0 & 0 & 0 & 0 & 0 & 0 & 0 & 0 \\ 0 & 0 & 0 & 0 & 0 & 0 & 0 & 0 \\ 0 & 0 & 0 & 0 & 0 & 0 & 0 & 0 \\ \frac{K_{T}}{m} & \frac{K_{T}}{m} & \frac{K_{T}}{m} & \frac{K_{T}}{m} & \frac{K_{T}}{m} & \frac{K_{T}}{m} & \frac{K_{T}}{m} & \frac{K_{T}}{m} \\ 0 & 0 & 0 & 0 & 0 & 0 & 0 & 0 \\ 0 & 0 & -\frac{dK_{T}}{I_{xx}} & -\frac{dK_{T}}{I_{xx}} & 0 & 0 & \frac{dK_{T}}{I_{xx}} & \frac{dK_{T}}{I_{xx}} \\ 0 & 0 & 0 & 0 & 0 & 0 & 0 & 0 \\ -\frac{dK_{T}}{I_{yy}} & -\frac{dK_{T}}{I_{yy}} & 0 & 0 & \frac{dK_{T}}{I_{yy}} & \frac{dK_{T}}{I_{yy}} \\ 0 & 0 & 0 & 0 & 0 & 0 & 0 & 0 \\ -\frac{K_{D}}{I_{zz}} & \frac{K_{D}}{I_{zz}} & -\frac{K_{D}}{I_{zz}} & \frac{K_{D}}{I_{zz}} & -\frac{K_{D}}{I_{zz}} & \frac{K_{D}}{I_{zz}} & -\frac{K_{D}}{I_{zz}} & \frac{K_{D}}{I_{zz}} 
    \end{bsmallmatrix}}^\text{B}
    \overbrace{
    \begin{bsmallmatrix}
    \Omega_{1}^{2} \\ \Omega_{2}^{2} \\ \Omega_{3}^{2} \\ \Omega_{4}^{2} \\ \Omega_{5}^{2} \\ \Omega_{6}^{2} \\ \Omega_{7}^{2} \\ \Omega_{8}^{2}
    \end{bsmallmatrix}}^\text{$u(t)$}
\end{eqnarray*}
\normalsize
Since the Mars coaxial quadrotor control system is designed for a position and heading controller, the output variables are $[x\ y\ z\ \psi]$ and the equation (\ref{ss2}) can be written as:
\small
\begin{equation*}
    \overbrace{
    \begin{bsmallmatrix}
    x \\ y \\ z \\ \psi
    \end{bsmallmatrix}}^\text{$Y(t)$}
    = 
    \overbrace{
    \begin{bsmallmatrix}
    1 & 0 & 0 & 0 & 0 & 0 & 0 & 0 & 0 & 0 & 0 & 0 \\ 0 & 0 & 1 & 0 & 0 & 0 & 0 & 0 & 0 & 0 & 0 & 0 \\ 0 & 0 & 0 & 0 & 1 & 0 & 0 & 0 & 0 & 0 & 0 & 0 \\ 0 & 0 & 0 & 0 & 0 & 0 & 0 & 0 & 0 & 0 & 1 & 0
    \end{bsmallmatrix}}^\text{C}
    \overbrace{
    \begin{bsmallmatrix}
    x \\ \dot{x} \\ y \\ \dot{y} \\ z \\ \dot{z} \\ \phi \\ \dot{\phi} \\ \theta \\ \dot{\theta} \\ \psi \\ \dot{\psi}
    \end{bsmallmatrix}}^\text{$x(t)$}
    + 
    \overbrace{
    \begin{bsmallmatrix}
    0 & 0 & 0 & 0 & 0 & 0 & 0 & 0 \\ 0 & 0 & 0 & 0 & 0 & 0 & 0 & 0 \\ 0 & 0 & 0 & 0 & 0 & 0 & 0 & 0 \\ 0 & 0 & 0 & 0 & 0 & 0 & 0 & 0
    \end{bsmallmatrix}}^\text{D}
    \overbrace{
    \begin{bsmallmatrix}
    \Omega_{1}^{2} \\ \Omega_{2}^{2} \\ \Omega_{3}^{2} \\ \Omega_{4}^{2} \\ \Omega_{5}^{2} \\ \Omega_{6}^{2} \\ \Omega_{7}^{2} \\ \Omega_{8}^{2}
    \end{bsmallmatrix}}^\text{$u(t)$}
\end{equation*}
\normalsize

\section{Model Predictive Control}
Model Predictive Control also known as Receding Horizon Control (RHC) uses the mathematical model of the system in order to solve a finite, moving horizon and closed loop optimal control problem~\cite{lopes2011model}. Thus, the MPC scheme is able to utilize the information about the current state of the system in order to predict future states and control inputs for the system~\cite{mayne2000constrained}. The main advantage of using the MPC scheme is the ability to take into account the physical limitations of the system, as well as the external noise. Due to these considerations in the design process, the controller is able to predict the future output of the system and in order to formulate an optimal control effort that brings the system to a desired state for a predefined reference trajectory~\cite{alexis2011model}. In the controller design the ability to introduce constrains allows the control inputs and output to be in a predefined bound for the desired performance. In general, the state and input constrains are set in relation to the application profile of the control scheme~\cite{8796236}. In the MPC design, the control signals need to be computed online for every sampling time, while considering the constrains at the same time. Forward shift operators are used in the conventional MPC design to optimize future control signals~\cite{chen2013cascaded}. For a complex dynamic system, a large prediction and control horizon is required, which results in increased online computations. There have been studies that propose methods to decrease the factors affecting the online computation as discussed in~\cite{wang2009model}. 

The optimal control problem is solved at each sampling time interval for the predefined prediction horizon. The first step of the solution of the optimization problem is applied to the system until the next sampling time~\cite{ru2017nonlinear}. This process can be repeated over and over in order to correctly compute the control inputs for the system to reach the desired state. The earlier discussed nonlinear mathematical model of the Mars coaxial quadrotor is linearized in order to design a linear MPC. Therefore, the equivalent discretized linear system for a sampling time $T_{s}$ can be described as:

\begin{equation*}
    \Delta x_{k+1} = \textbf{A}\Delta x_{k} + \textbf{B}\Delta u_{k}
\end{equation*}
\begin{equation*}
    \Delta y_{k} = \textbf{C}\Delta x_{k} 
\end{equation*}

where, $\Delta x_{k}=x_{k}-x_{T}$ and $\Delta u_{k}=u_{k}-u_{T}$ and $k$ is the current sample. In the proposed model, the state matrix $\mathbf{A \in \mathbb{R}}^{12\times 12}$, the input matrix $\mathbf{B \in \mathbb{R}}^{12\times 8}$, the $\mathbf{C \in \mathbb{R}}^{4\times 12}$ output matrix are as described in the state space representation of the linearized system. The reference point is considered at near hover condition, while the nominal control input will be $u = \frac{mg}{8}[1, 1, \dots, 1]_{8 \times 1}$. Based on the linear state space model, the designed controller predicts the future states as a function of current state and future control inputs. Therefore for $i = 1,2,3,..., N$  
\begin{equation*}
    \Delta x_{k+i+1} = \textbf{A}\Delta x_{k+i} + \textbf{B}\Delta u_{k+i}
\end{equation*}
\begin{equation*}
    \Delta y_{k+i} = \textbf{C}\Delta x_{k+i} + \textbf{D}\Delta u_{k+i}
\end{equation*}
\begin{equation*}
    \Delta X_{k} = \textbf{G}\Delta x_{k} + \textbf{H}\Delta u_{k}
\end{equation*}
\begin{equation}
    \Delta Y_{k} = \Bar{\textbf{C}}\Delta X_{k} + \Bar{\textbf{D}}\Delta U_{k}
    \label{pr1}
\end{equation}

\begin{figure}[h!]
  \centering
    \includegraphics[width=1\linewidth]{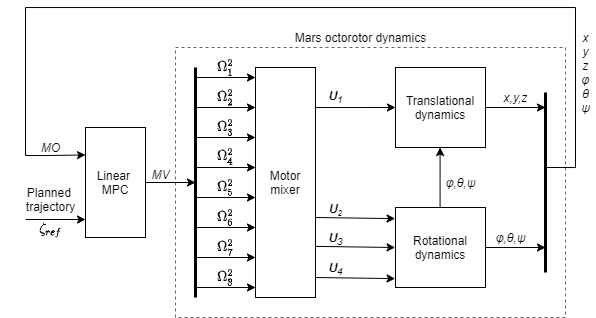}
    \caption{Control loop of Mars octorotor}
  \label{fig:mpcsblockdia}
\end{figure}

Equation (\ref{pr1}) becomes the prediction equation in which,
\small
\begin{equation*}
    \Delta X_{k} = 
    \begin{bsmallmatrix}
    \Delta x_{k} \\ \Delta x_{k+1} \\ \Delta x_{k+2} \\ \vdots \\ \Delta x_{k+N-1}
    \end{bsmallmatrix} ; 
    \Delta U_{k} = 
    \begin{bsmallmatrix}
    \Delta u_{k} \\ \Delta u_{k+1} \\ \Delta u_{k+2} \\ \vdots \\ \Delta u_{k+N-1}
    \end{bsmallmatrix} ; 
    \Delta Y_{k} = 
    \begin{bsmallmatrix}
    \Delta y_{k} \\ \Delta y_{k+1} \\ \Delta y_{k+2} \\ \vdots \\ \Delta y_{k+N-1}
    \end{bsmallmatrix}
\end{equation*}
\begin{equation*}
    \textbf{G} = 
    \begin{bsmallmatrix}
    I \\ A \\ A^{2} \\ \vdots \\ A^{N-1}
    \end{bsmallmatrix} ; 
    \textbf{H} = 
    \begin{bsmallmatrix}
    0 &\ &\ &\ \\ B & 0 &\ &\ \\ AB & B & 0 &\ \\ \vdots & \vdots & \ddots & \ddots &\ \\ A^{N-2}B & A^{N-3}B & \dots & B & 0
    \end{bsmallmatrix}
\end{equation*}
\begin{equation*}
    \Bar{\textbf{C}} = 
    \begin{bsmallmatrix}
    C &\ &\ &\ &\ \\ \ & C &\ &\ &\ \\ \ &\ & C &\ &\ \\ \ &\ &\ &\ &\ \ddots &\ \\ \ &\ &\ &\ &\ & C 
    \end{bsmallmatrix} ; 
    \Bar{\textbf{D}} = 
    \begin{bsmallmatrix}
    D &\ &\ &\ &\ \\ \ & D &\ &\ &\ \\ \ &\ & D &\ &\ \\ \ &\ &\ &\ &\ \ddots &\ \\ \ &\ &\ &\ &\ & D 
    \end{bsmallmatrix}
\end{equation*}
\normalsize
In order to choose the optimal control input at each sampling time interval, a cost function formulation is considered for the MPC design. The aim of the cost function formulation is to minimize the control effort, while driving the predict outputs to a reference trajectory. In the proposed controller design the cost function penalizes the error in the reference trajectory and the current output state, as well as the control inputs.
\begin{align*}
    Q(\Delta x_{k},\Delta U_{k}) = (\Delta x^{ref}_{k}-\Delta x_{k})^T \Bar{M_{x}}(\Delta x^{ref}_{k}-\Delta x_{k}) + \\ 
    (\Delta u^{ref}_{k}-\Delta u_{k})^T \Bar{M_{u}}(\Delta u^{ref}_{k}-\Delta u_{k}) + \\
    (\Delta u_{k}-\Delta u_{k-1})^T \Bar{M_{\Delta u}}(\Delta u{k}-\Delta u_{k-1})
\end{align*}

where $\Bar{M_{x}}$, $\Bar{M_{u}}$ and $\Bar{M_{\Delta u}}$ are the weighing matrices for the states, the control inputs and the rate of change in the control inputs. The linear MPC is validated using the linearized plant model of the Mars coaxial quadrotor mathematical model. The linearized model uses the state space form of linear dynamics, where a near hover condition is considered as the reference point. Once the MPC design is validated, the same controller is used with the non linear model of Mars coaxial quadrotor. The closed loop control architecture is presented in~\autoref{fig:mpcsblockdia}. The term $\zeta_{ref} = [x\ y\ z\ \psi]^T$ represents reference trajectory to follow.

\subsection{Closed loop Simulation}

\begin{figure}[h!]
    \centering
    \subfigure[X position response]
    {
        \includegraphics[width=1\linewidth]{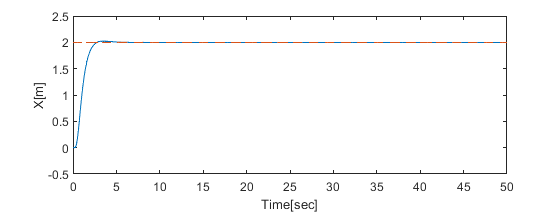}
    }
    \subfigure[Y position response]
    {
        \includegraphics[width=1\linewidth]{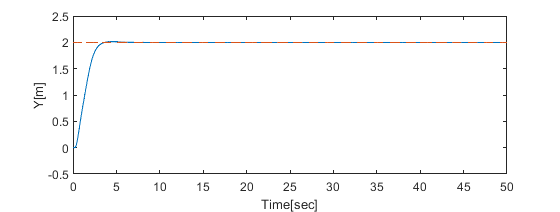}
    }
    \subfigure[Z position response]
    {
        \includegraphics[width=1\linewidth]{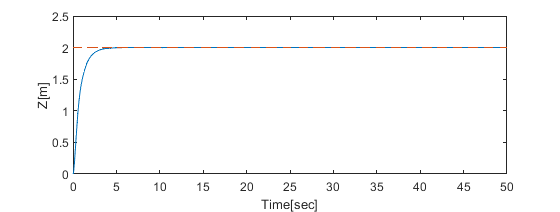}
    }
    \caption{Position response}
    \label{posrespons}
\end{figure}

The proposed design of the linear MPC uses a nonlinear model of the Mars coaxial quadrotor for the presented simulation results in this Section. The position controller responses are recorded in order to validate the MPC design. A constant reference in $x,\ y,\ $ and $z$ are given and the MPC responses for position tracking are presented in \autoref{posrespons}.

\begin{figure}[h!]
    \centering
    \subfigure[Square trajectory tracking using MP-controller]
    {
        \includegraphics[width=0.8\linewidth]{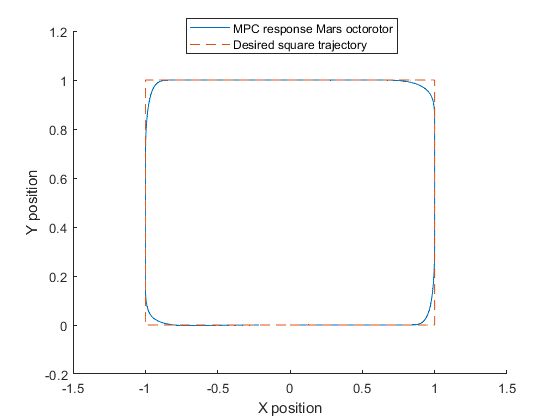}
        \label{sqr}
    }
    \subfigure[Helix trajectory tracking using MP=controller]
    {
        \includegraphics[width=0.8\linewidth]{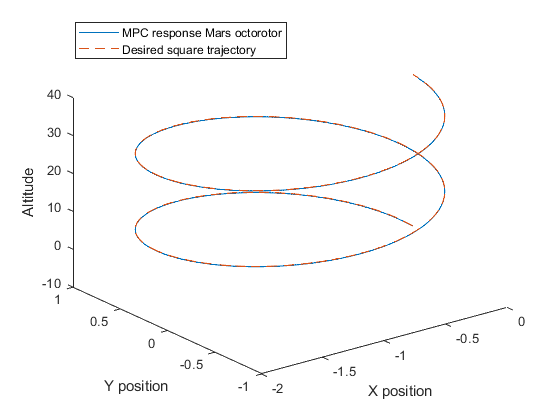}
    }
    \caption{Trajectory tracking with MPC}
    \label{trajectories}
\end{figure}

\begin{figure}[h!]
    \centering
    \subfigure[X position during helix trajectory tracking]
    {
        \includegraphics[width=0.8\linewidth]{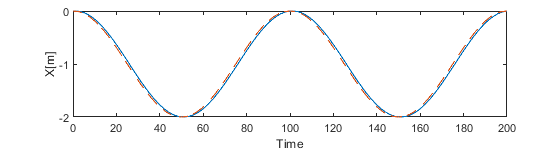}
    }
    \subfigure[Y position during helix trajectory tracking]
    {
        \includegraphics[width=0.8\linewidth]{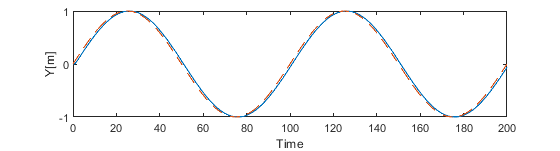}
    }
    \subfigure[Z position during helix trajectory tracking]
    {
        \includegraphics[width=0.8\linewidth]{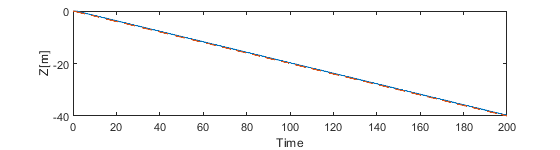}
    }
    \caption{Position response for helix trajectory}
    \label{posresponshelix}
\end{figure}

The trajectory following responses are presented in \autoref{trajectories} and the position responses during a reference helix trajectory tracking are presented in \autoref{posresponshelix}. For the first simulated helix trajectory, the reference parameters are defined as:
\begin{equation*}
    x^{ref} = 1cos(0.02\pi t)\ m ;\ y^{ref} = 1sin(0.02\pi t)\ m
\end{equation*}
\begin{equation*}
    z^{ref} = 0.1t\ m ;\ \psi^{ref} = 0\ rad
\end{equation*}

\begin{figure}[h!]
    \centering
    \includegraphics[width=0.8\linewidth]{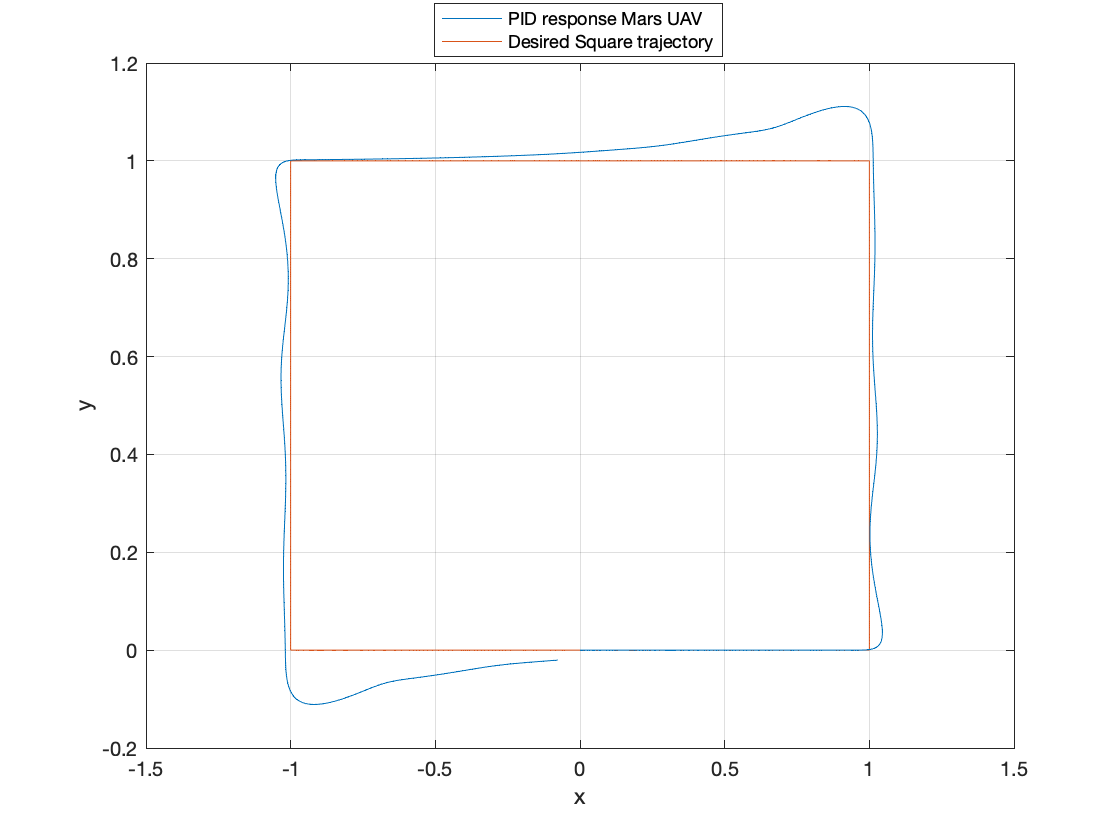}
    \caption{Trajectory follow response based on PID controller}
    \label{pid}
\end{figure}

From \autoref{sqr} it can be noted that at the sharp turn in the reference square trajectory, the designed MPC predicts control inputs in such a way that the coaxial quadrotor does not over shoot, while maintaining a minimal control effort. This is one of the attractive feature of the Model Predictive Control scheme when compared against the PID control scheme. For comparison purpose, the PID controller also uses the exact same nonlinear model of Mars coaxial quadrotor. The trajectory tracking using a PID controller is shown in~\autoref{pid}. As mentioned before, at the sharp turn in the reference square trajectory, the PID controller performs poorly, when compared to the MPC response, in terms of overshoot and control effort. The same can be seen clearly in~\autoref{pid}.

\section{Conclusions}

In this article the design and control oriented modelling of a Mars coaxial quadrotor was presented. Based on the state of the art on an  aerial vehicle flight in Mars atmosphere, our proposed design allows for a bigger payload for exploration and mapping purposes. In terms of increased payload, the proposed model outperforms the state of the art because it can accommodate larger solar arrays that can be extended or retracted. Mars has low gravity but also thin atmosphere therefore, the proposed design is optimized in terms of revolutions speed and rotor size for a realistic scenario. The results from the flow simulation validate the large size of the vehicle, in terms of the thrust produced by pairs of coaxial rotors. The Mars coaxial quadrotor uses a sufficient linear MPC based control architecture. The MPC based control compliments the overall autonomy of the vehicle when it is used for the exploration purpose. Based on the linear MPC, the proposed actuator based model of the coaxial quadrotor produces an optimal control inputs such that the vehicle avoids aggressive maneuvers in order to maintain a stable flight and to save power as well. This has been validated by following a square trajectory and with a comparison against a PID based model in different simulated trajectories.


\bibliographystyle{IEEEtran}
\bibliography{mybib}


\end{document}